%% file: sample-acmsmall.tex
\documentclass[acmsmall]{acmart}

\usepackage{multirow}
\usepackage{longtable}
\usepackage{amsmath}
\usepackage{footnote}
\makesavenoteenv{tabular}
\usepackage{tablefootnote}
\usepackage{array}
\newcolumntype{P}[1]{>{\centering\arraybackslash}p{#1}}

\usepackage{lscape}
\usepackage{diagbox}
\usepackage[para]{footmisc}
\usepackage[para]{footmisc}

\usepackage{xcolor}
\definecolor{softred}{rgb}{1.0, 0.13, 0.32}
\definecolor{softblue}{rgb}{0.0, 0.5, 1.0}

\definecolor{light-gray}{HTML}{E5E4E2}
\definecolor{light-cyan}{HTML}{E0FFFF}

\usepackage{caption}
\usepackage{subcaption}
\usepackage{enumitem}

\AtBeginDocument{%
  }

\definecolor{RoyalBlue}{rgb}{0, 0, 0}
\definecolor{purple}{rgb}{0, 0, 0}
\definecolor{forestgreen}{rgb}{0, 0, 0}
\newcommand*{\editing}{\textcolor{RoyalBlue}}
\newcommand*{\moving}{\textcolor{forestgreen}}
\newcommand*{\minorediting}{\textcolor{purple}}

\begin{document}

\title[An Audit of Bengali Sentiment Analysis Tools and Their Identity-based Biases]{The ``Colonial Impulse" of Natural Language Processing: An Audit of Bengali Sentiment Analysis Tools and Their Identity-based Biases}

\author{Dipto Das}
\email{dipto.das@colorado.edu}
\affiliation{%
  \institution{Department of Information Science, University of Colorado Boulder}
  \city{Boulder}
  \state{Colorado}
  \country{United States}
}
\author{Shion Guha}
\email{shion.guha@utoronto.ca}
\affiliation{
  \institution{Faculty of Information, University of Toronto}
  \city{Toronto}
  \state{Ontario}
  \country{Canada}
}
\author{Jed Brubaker}
\email{jed.brubaker@colorado.edu}
\affiliation{
  \institution{Department of Information Science, University of Colorado Boulder}
  \city{Boulder}
  \state{Colorado}
  \country{United States}
}
\author{Bryan Semaan}
\email{bryan.semaan@colorado.edu}
\affiliation{
  \institution{Department of Information Science, University of Colorado Boulder}
  \city{Boulder}
  \state{Colorado}
  \country{United States}
}

\begin{abstract}
  While colonization has sociohistorically impacted people's identities across various dimensions, those colonial values and biases continue to be perpetuated by sociotechnical systems. One category of sociotechnical systems--sentiment analysis tools--can also perpetuate colonial values and bias, yet less attention has been paid to how such tools may be complicit in perpetuating coloniality, although they are often used to guide various practices (e.g., content moderation). In this paper, we explore potential bias in sentiment analysis tools in the context of Bengali communities who have experienced and continue to experience the impacts of colonialism. Drawing on identity categories most impacted by colonialism amongst local Bengali communities, we focused our analytic attention on gender, religion, and nationality. We conducted an algorithmic audit of all sentiment analysis tools for Bengali, available on the Python package index (PyPI) and GitHub. Despite similar semantic content and structure, our analyses showed that in addition to inconsistencies in output from different tools, Bengali sentiment analysis tools exhibit bias between different identity categories and respond differently to different ways of identity expression. Connecting our findings with colonially shaped sociocultural structures of Bengali communities, we discuss the implications of downstream bias of sentiment analysis tools.
\end{abstract}

\maketitle

\input{sections/introduction}
\input{sections/literature_review}
\input{sections/methods}
\input{sections/results}
\input{sections/discussion}
\input{sections/conclusion}

\section*{Acknowledgements}
We thank the tool developers who responded to our communication to provide their demographic information and answer our queries about reproducing their tools.

\bibliographystyle{ACM-Reference-Format}
\bibliography{sample}


\end{document}

%% file: sections/introduction.tex
\section{Introduction}
Natural language processing (NLP) enables computers to ``understand," ``interpret," and ``generate" language. One kind of NLP is centered around analyzing "sentiment," which is the process of determining the emotional tone expressed in text data. Though it is widely used in computational linguistics, HCI researchers have critiqued this approach. Sentiment analysis, which seeks to assign subjectivity or polarity scores (usually within standardized scales) or nominal sentiment categories (e.g., positive, negative, neutral),  becomes an exercise of quantifying and categorizing complex human language and emotion. However, researchers have highlighted how the process of sorting and categorization are political and reductionist and can perpetuate inequality~\cite{bowker2000sorting, dourish2012ubicomp}. When such processes are used by computing systems to interpret and analyze human language, their operations and outcomes often include social and technical biases~\cite{friedman1996bias}.

Critical algorithmic studies scholars defined bias as when computer systems consistently and unfairly discriminate against certain individuals or groups in favor of others~\cite{friedman1996bias}. Social power structures, global resource availability, and biases can manifest in various ways through computing systems. Especially in NLP, there is an incredible disparity in research and resources available across various languages. Joshi and colleagues identified 0.28\% of languages as ``the winners" and 88.38\% of languages as the ones ``left behind"~\cite{joshi-etal-2020-state}. For example, in comparing language resources across English and Bengali, they found that although English and Bengali have comparable numbers of speakers~\cite{lane2023most}, English has hundreds of times higher visibility than Bengali in terms of resources on Linguistic Data Consortium, Wikipedia, and publication venues like Language Resources and Evaluation~\cite{joshi-etal-2020-state}. Besides the resource disparities across languages, attention to how bias works in non-English systems has not been explored. Bias can work differently in different languages and cultures. Imposing Euro-centric (e.g., English) language technologies on diverse user communities without considering their cultural and historical contexts can have deleterious impacts. Applying NLP tools designed in the West to other language and cultural traditions can undermine ``safety measures" (e.g., in content moderation)~\cite{nicholas2023lost, nicholas2023re} and impose Western values and perspectives. Since artificial intelligence (AI)-based technologies disproportionately harm marginalized communities like non-native English speakers~\cite{reventlow2021artificial, akselrod2021artificial}, researchers have called for increased focus on non-English NLP studies~\cite{nicholas2023re, argoub2021nlp}.

\editing{In this paper, we employ a sociotechnical approach to our exploration of NLP tools. Here, when using the phrase ``sociotechnical systems," we are not referring to a specific tool or set of technologies/tools but all technology that shapes and is shaped by human interaction~\cite{sawyer2014sociotechnical}. We know from prior work that artifacts like algorithms and machine learning (ML) technologies are political and are shaped by societal norms as well as the individual or developer group's politics within which they are designed~\cite{winner2017artifacts, scheuerman2021datasets}. Sentiment analysis tools, in particular, are sociotechnical in how they shape and are shaped by human interaction. On the one hand, people develop these tools, and user interaction data is often used to train these tools, which shapes their outputs. On the other hand, let's draw on the example of content moderation, where outputs from these tools are used in decision-making (e.g., ~\cite{sun2022design, vaidya2021conceptualizing}). When used in such moderated spaces, they shape users' social interactions. Altogether, these interdependencies demonstrate sentiment analysis tools' (a) mutual constitution of social and technological factors, (b) contextual embeddedness of this mutuality (e.g., in various sociocultural settings), and (c) collective action of tool developers and users--three elements of the sociotechnical premise as outlined by Sawyer and Mohammad~\cite{sawyer2014sociotechnical}. Therefore, NLP tools like those for sentiment analysis are sociotechnical systems ~\cite{venkit2023sentiment}.}

As people continue to adopt computational linguistic systems, the possibility for the propagation of harmful decisions made with their assistance can have downstream effects--consequences that are experienced at a later stage. \minorediting{Therefore}, it is incredibly important to understand the application of NLP in non-Western settings. To address these myriad concerns, our research foregrounds non-English NLP research, particularly \editing{sentiment analysis} in the Bengali \moving{language\footnote{``Bangla" is the endonym for the Bengali language, used by native speakers, while ``Bengali" is the exonym popularly used by people from other linguistic and cultural backgrounds to refer to the same language and its speakers. Bengali ranks as the sixth most widely spoken native language (with around 259.89 million speakers) and the seventh most spoken language overall (with approximately 267.76 million speakers) globally~\cite{das2021jol}.}}, from the perspective of fairness and bias. We investigated how Bengali sentiment analysis (BSA) tools assess specific identities, explore differences in their responses for explicit and implicit identity expressions, and examine potential biases across different identity categories and the relationship between bias and tool developer demographics. The Bengali language is natively spoken by the Bengali people (endonym Bangali), who are native to the Bengal region in South Asia that constitutes present-day Bangladesh and the West Bengal state of India. Historically, these communities were significantly impacted by prolonged British and Pakistani colonization~\cite{das2022collaborative, ali1971bangla}--the practices of foreign powers migrating to and altering the social structures of local communities~\cite{loomba2002colonialism}.

While colonization impacted communities globally, postcolonial computing scholars argue that sociotechnical systems \moving{continue to reinforce colonial values and hierarchies, especially in the Global South contexts~\cite{irani2010postcolonial, dourish2012ubicomp}}. \minorediting{According to Dourish and Mainwaring, these systems are shaped by and through a ``colonial impulse"\editing{--``a series of considerations" that relies on and reinforces universality, reductionist representation, and colonial hierarchies and politics.}~\cite{dourish2012ubicomp}}. When computer systems embody preexisting biases, they can discriminate against populations often based on identity~\cite{friedman1996bias}. Identity is a person's understanding of who they are and how they want others to see them as social and physical beings~\cite{goffman1978presentation, gecas1982self, erikson1968identity}. It is often perceived through one's race, gender, nationality, religion, etc.~\cite{tajfel1974social}. \minorediting{Similar to how the identities of} Bengali communities \minorediting{have been} impacted \minorediting{by colonialism across various dimensions} \moving{(as elaborated in section~\ref{sec:literature_review})}, \editing{in studying the \textit{colonial impulse} of sentiment analysis tools, we explore whether and how these tools reduce Bengali identities to only religion or nationality, reinforce ``traditional" views on gender, and reanimate colonial hierarchies and prejudices by regarding certain identities as more positive or negative.}

In this paper, we seek to understand whether and how BSA tools reanimate colonially shaped social biases across these identity dimensions by asking the following research questions:

\begin{itemize}[leftmargin=*]
    \item[] \minorediting{\textbf{RQ1.a:}} How do different tools differ in assigning sentiment scores to a particular identity?
    \item[] \minorediting{\textbf{RQ1.b:}} How do scores differ between explicit and implicit expressions of identity?
    \item[] \minorediting{\textbf{RQ2.a:}} Do BSA tools show biases across gender, religious, and national identity categories?
    \item[] \minorediting{\textbf{RQ2.b:}} What is the relationship between \minorediting{tools'} bias and \minorediting{developers' demographic backgrounds?}
\end{itemize}

To answer these questions, we conducted an algorithmic audit of BSA tools available on PyPI and GitHub. Looking at different genders, religions, and nationalities, we found that \minorediting{different BSA tools assign significantly different sentiment scores for identical sentences expressing a particular identity.} \minorediting{In particular, BSA tools often rate an explicit expression of Bengali identity based on nationality more negatively than when the same identity is expressed implicitly.} \minorediting{We also found the majority of tools to be biased. \editing{Among the 13 tools we audited, 38\% and 30\% are respectively biased toward female and male gender identities, 30\% and 38\% are biased across religious (e.g., Hindu and Muslim), and 77\% and 15\% were biased across nationality-based identities (e.g., Bangladeshis and Indians)}--reanimating the colonial hierarchies.} Though we found a digital divide among diverse Bengali communities in developing language technologies, \minorediting{our analysis did not suggest that the demographics of the developers conclusively affect the bias within sentiment tools}. Taken together, our work highlights how BSA tools exhibit a ``colonial impulse." We discuss the downstream implications of using available BSA tools and provide recommendations for future research.

%% file: sections/literature_review.tex
\section{Literature Review}\label{sec:literature_review}
\subsection{How Colonialism Impacted Social Identities in Bengali Communities}
While identity is often construed as an individuated concept, identities are often influenced by people's cultural background and social interactions~\cite{butler2011gender, anderson2006imagined}. Thus, various social identities emerge centered around people's perceived membership in different groups~\cite{tajfel1974social}. In this view, people's identities are defined across various \textbf{dimensions}, such as race, ethnicity, gender, sexual orientation, religion, nationality, and caste. Within each dimension \minorediting{(e.g., religion)}, people can identify with different \textbf{categories} \minorediting{(e.g., Christian)}~\cite{mccall2005complexity}. Importantly, people's identities across various dimensions interconnect and overlap, and the consequent intersectional identities collectively shape their unique experiences, social position\minorediting{, and systemic privilege~\cite{collins2022black, crenshaw2013demarginalizing}}. This is best illustrated through how marginalization--the process wherein people are pushed to the boundary of society and denied agency and voice based on their intersectional social identities--is normalized through cultural hegemony \cite{collins2022black, crenshaw2013demarginalizing}. Cultural hegemony is a system of ideas, practices, and social relationships embedded within private and institutional domains as a mechanism of power and control. Through cultural hegemony, people are categorized as a mechanism of power where some identities are considered ``normative" while others are considered non-normative. In other words, people experience everyday harm and are marginalized by virtue of being born Black, Queer, or into a lower Caste.

A global practice that shaped and continues to shape the hegemonic structures of society and, in turn, people's everyday experiences is coloniality. \minorediting{While colonization has deeply impacted people's identity, coloniality refers to its enduring and pervasive effects on the local and indigenous communities even after the direct colonial rule has ended~\cite{mignolo2007delinking}. These continue to perpetuate colonial structures and social, economic, political, and cultural dynamics}. Among other dimensions of identity, European colonialism imposed its conceptualization of gender on many indigenous communities~\cite{lugones2007heterosexualism}. Scholars have studied colonized Bengali societies to understand the complex relationship between colonialism and gender~\cite{sinha2017colonial, dimeo2002colonial}. British colonization, they argue, produced a particular kind of masculine identity, wherein the ``manly Englishman" was contrasted with the stereotyped ``effeminate Bengali" in order to justify British rule and denigrate Bengali culture~\cite{sinha2017colonial}. Such colonial masculinity had profound impacts on gender and ethnic relations. This view led to the stereotyped views of Bengali men in colonial India~\cite{dimeo2002colonial, rand2006martial} and the reinforcement of ``traditional gender roles" in Bengal~\cite{sinha2016study}. This minimized women's sociopolitical participation \minorediting{and voices~\cite{spivak2023can}}.

The imposition of European standards also distorted people's religious values and perceptions of the Indian subcontinent. Scholars have attributed the rise of religious extremism and the violence against minorities in the region to colonial values and divide-and-rule practices~\cite{nandy1989intimate, das2006life}. They argue that religion-based nationalism is a reactive ideology that emerged in response to the challenges posed by colonialism and the West, where local people have adopted many ideas and practices of Abrahamic religions, such as the emphasis on a single, monolithic God\minorediting{~\cite[p. 24]{nandy1989intimate}} and the belief in a chosen people\minorediting{~\cite[p. 101]{nandy1989intimate}}. Especially due to cultural assimilation--the idea that colonizers' culture is superior to that of the native communities~\cite{fanon2008black} and cultural genocide--the destruction and theft of cultural sites and artifacts~\cite{van2004rethinking}, as the colonized subjects were denied the opportunities to explore, understand, and practice their own culture, local and native communities' self-perception regarding religion changed. Moreover, the British colonizers amplified, exploited, and institutionalized local communities' religious differences and divisions~\cite{chatterjee1993nation}.

Across the world, colonizers introduced classifications to partition different nation-states based on their own perceptions of nationhood and societal groupings of the native communities (e.g., two-nation theory in India-Pakistan)~\cite{greenberg2005generations}. Such outlooks disregarded the latter's intricate self-perceptions and interconnectedness~\cite{chatterjee1993nation}. \minorediting{Before} their departure in 1947, British colonizers partitioned the Indian subcontinent, prioritizing religion as the only dimension of people's collective identity. In the context of Bengal, West Bengal, with its upper-caste Hindu majority, was annexed to India, while East Bengal, characterized by a Muslim and underprivileged-caste Hindu majority, became a part of Pakistan~\cite{sen2018decline}. This \minorediting{displaced} millions of Bengalis as refugees across the India-Pakistan border~\cite{pandey2001remembering} and marginalized the Bengali people under Pakistani subjugation~\cite{ali1971bangla} as the \minorediting{long} geographic distance and myriad cultural differences between Pakistan and East Bengal were overlooked in this colonially imposed idea of nationality. Eventually, in 1971, East Bengal gained independence from Pakistan and \minorediting{formed} Bangladesh based on people's ethnolinguistic identity.

Overall, among myriad dimensions of marginalization, colonization crucially impacted the expression of social identities in the context of Bengali communities by impacting their perception of gender roles of men and women, the religious division of Hindus and Muslims, and the socio-economic structures and political consciousness culminating in Bengali communities assuming different nationalities (e.g., Bangladeshi and Indian). 

\subsection{Expressions of Social Identity through Language and Technology}
This coloniality has continued to shape people's everyday experiences and, on a deeper level, mediate how they express their social identities. One can express one's social identity both explicitly and implicitly. Explicit expressions of identity refer to deliberate and direct ways individuals communicate and assert their affiliations, characteristics, and beliefs. For example, mentioning one's nationality and political views or openly discussing one's religious beliefs are examples of explicit expressions of identity~\cite{tajfel1974social}. Meanwhile, implicit expressions of identity include subtle and indirect ways in which identity is communicated or inferred from a person's actions, behaviors, choices, and interactions~\cite{turner1987rediscovering} and are bound up with cultural norms, societal expectations, and institutionalized practices~\cite{butler2011gender, hovy2020you}. For example, how one speaks, the words they use, or their hobbies can implicitly give insights about one's identity. While people's social identities can be communicated implicitly through different speech acts and non-verbal acts, this paper focuses on linguistic expressions of various identity categories through writing. Particularly, we considered how different gender, religion, and nationality-based identities are expressed explicitly and implicitly in Bengali texts.

Cultural-linguistic scholars have detailed how languages are often standardized differently in different countries  (e.g., English in England vs. the United States; German in Germany vs. Austria)~\cite{brown2020perspectives}. These geo-cultural variations, often referred to as dialects, operate as important signs and implicit expressions of cultural identity~\cite{falck2012dialects, hershcovich2022challenges}. In Bengali, the two main dialects are \textit{Bangal} and \textit{Ghoti}, which are spoken in East Bengal (Bangladesh) and West Bengal (in India), respectively ~\cite{das2023toward}. These variations of the Bengali language manifest both phonologically and textually~\cite{kibria2022bangladeshi, bhasa2001praci} and use different colloquial vocabularies in written texts for the same everyday objects. For example, Bangladeshi and Indian Bengalis respectively use the words \textit{jol} and \textit{pani} to mean ``water." Consistently using vocabulary from either the Bangal or Ghoti dialects can implicitly express a Bengali person's national identity without any explicit mention. Similarly, Bengali textual communication often implies the gender and religious identities of the people it describes. While \minorediting{in Bengali, unlike many other Indo-European languages, gender does not change the choice of pronouns (as in English) and verbs (as in Hindi and Urdu)}~\cite{bhattacharjee2021banglabert}, culturally, most names and kinship terms are gender-specific with some exceptions~\cite{dil1972hindu}. Moreover, commonly used kinship terms, names, and commonly used vocabularies often implicitly indicate one's membership or being born into either Hindu or Muslim communities~\cite{dil1972hindu, das2023toward}. For example, while Bengali Hindus often draw inspiration from Demigods' names and characters in legends for their personal names and commonly tend to use Bengali words derived from Sanskrit, in Bengali Muslim communities being named after Prophets, Caliphs, and Mughal emperors and the vernacular use of Perso-Arabic words are widely popular~\cite{dil1972hindu}. Thus, written Bengali communication can lead to the inference of one's gender, religion, and nationality-based identities.

As the colonizers invented categorization and classifications by viewing and interpreting cultures, societies, and people from non-Western locations in a stereotyped and exoticized manner~\cite{said2014orientalism}, hierarchies among these artificial categories have been established and embedded within colonized societies~\cite{fanon2008black, das2022collaborative}. Broadly, these experiences included everything from colonially shaped racism (a belief in certain racial groups' inherent superiority or inferiority) to colorism (favoring lighter skin tones over darker ones within a single racial group). With respect to how people express their social identities through written language, the influence and affluence of West Bengal's upper-caste Hindu landlords and elites, who predominantly spoke the \textit{Ghoti} dialect, led to the establishment of their dialect as the institutional and ``normative" standard for the Bengali language during the introduction of printing presses in the region~\cite{chatterjee1993nation}. In contrast, the \textit{Bangal} dialect became associated with East Bengal's agrarian socioeconomic system and refugees due to mass migrations following the colonial partition and a means of Muslim and underprivileged caste Hindus' social harassment~\cite{das2021jol, ghoshal2021mirroring}. Through coloniality, these impacts on identity, such as sociolects (dialects of particular social classes~\cite{mccormack2011hexagonal}) and colonial ontologies and epistemologies--the ways of being and knowing--are embedded within the world structures at regional and global scales and continued across generations through various artifacts, media, and technology~\cite{ali2016brief, banerjee2015more}.

This leads to critical and important questions: Are sociotechnical systems ``mindful" of such sociocultural and historical complexities that shape people's identities? How are identities translated into ``something a microchip can understand"~\cite{rudder2013inside}?

\subsection{Algorithmic Bias Deconstruction in Computing Systems}
To better interrogate these questions, we draw on postcolonial computing scholarship. Broadly construed, postcolonial and decolonial scholars have worked to highlight the ``colonial impulse" of technology~\cite{irani2010postcolonial, dourish2012ubicomp}. \minorediting{Dourish and Mainwaring identified notions that undergird both colonial narratives and computing systems, such as belief in universality, reliance on reductive representation, and comparative evaluation of different sociocultural identities~\cite{dourish2012ubicomp}}. While prior critical HCI scholarship has studied the design and development of ubiquitous computing~\cite{dourish2012ubicomp} and computer vision~\cite{scheuerman2021auto} from postcolonial and decolonial perspectives, in this paper, we seek to understand how BSA tools reanimate social biases based on identities in previously colonized communities.

Computing systems construct people's algorithmic identities--how digital technologies and algorithms construct and represent individuals' identities through data-driven processes~\cite{cheney2017we}. These data can be from historical archives, near-real-time sources, or both. Since historical archives often reflect colonial \minorediting{ontologies and hierarchies}~\cite{taylor2003archive}, when used to inform computing systems like algorithms, they can inadvertently perpetuate these colonial \minorediting{values}~\cite{buolamwini2018gender}. Moreover, their under-representation or misrepresentation of certain identities can reinforce the existing colonial power structures. Even near-real-time data \minorediting{being interpreted through colonial taxonomies assign people to hierarchized categories across race, gender, or nationality}~\cite{cheney2017we}. Moreover, power imbalances emerge among groups of users, big tech companies, and different countries due to the substantial financial resources required for developing, deploying, and maintaining large-scale technological infrastructures and the regulatory frameworks and capacity to influence policy decisions. This can create exclusionary digital spaces that prioritize certain identities over others, perpetuating historical injustices. Therefore, scholars have described \minorediting{sociotechnical systems' approaches to conceptualizing people without considering social contexts as ``colonial impulses"~\cite{dourish2012ubicomp}}.

\minorediting{Sociotechnical systems, broadly construed, reanimate and reinforce existing societal power structures; they are likely to discriminate~\cite{benjamin2019race, saetra2021ai}. Scholars have explored how systems like facial recognition, predictive policing, hiring algorithms, facial beauty apps, recommendation systems, and standardized tests exhibit biases~\cite{benjamin2019race, cheney2017we, broussard2019artificial}. More specific to AI, beyond the biases that originate from individuals having significant input into the design of an AI system, biases also manifest from social institutions, practices, and values~\cite{ehsan2020human}}. Bias could also arise from technical constraints (e.g., while making qualitative human constructs quantitatively amenable to computers~\cite{dourish2012ubicomp}) as well as based on the context of use (e.g., users having different values from the system or dataset developers\minorediting{~\cite{sen2015turkers, ehsan2020human})}. \editing{AI systems' reductionist representations rely on codified stereotypes~\cite{benjamin2019race} and induce essentialization of certain identities~\cite{hanna2020towards}, \minorediting{which Scheuerman et al. in the case of computer vision (CV) characterized as an ``extended colonial project"~\cite{scheuerman2021auto}}}. Researchers in CHI and adjacent fields have recently been studying the biases and fairness of systems reliant on ML, NLP, and CV~\cite{mehrabi2021survey, blodgett2020language, scheuerman2020we}. Many of them proposed and used ``algorithmic audit" as a way to evaluate sociotechnical systems for fairness and detect their discrimination and biases~\cite{metaxa2021auditing}.

Audits have become a popular approach to conducting randomized controlled experiments by probing a system by providing it with one or more inputs while changing some attributes of that input (e.g., race, gender) in environments different from the system's development~\cite{metaxa2021auditing}. For example, Bertrand and Mullainathan's classic audit study~\cite{bertrand2004emily} tested for racial discrimination in hiring, specifically in reviewing resumes, created and submitted fictitious resumes with similar qualifications bearing white-sounding or Black-sounding names to job postings in many companies and industries and quantified the frequency at which those imaginary job seekers received interview callback responses. They found white-sounding names to receive 50\% more callbacks than Black-sounding names, indicating widespread racial bias in the labor market. Algorithm audits particularly examine algorithmic systems and content~\cite{sandvig2014auditing}. 

While some studies have delved into codes of open-source algorithms to study structural biases~\cite{johnson2017effect}, given that many algorithms we use are proprietary and like ``black boxes", algorithmic audits seek to decipher algorithms by interpreting output while varying inputs~\cite{diaz2018addressing, metaxa2021auditing}. This differs from other tests popularly used in computing and HCI literature. For example, unlike other common experiments in HCI, such as A/B tests in which the subject of the study is the users, in algorithmic audit, the subject of study is the system itself~\cite{metaxa2021auditing}. Algorithm audits are also different from other types of system testing due to their broader scope, resulting in systematic evaluations rather than binary pass/fail conclusions for individual test cases. Moreover, audits are purposefully intended to be external evaluations based only on outputs, without insider knowledge of the system or algorithm being studied~\cite{metaxa2021auditing}. Traditionally, querying an algorithm with a wide range of inputs and statistically comparing the corresponding results has been one of the most effective ways for algorithmic audits~\cite{metaxa2021auditing, sweeney2013discrimination-cacm}. Seminal work by Sweeney~\cite{sweeney2013discrimination-cacm, sweeney2013discrimination-queue} queried the Google Search algorithm with Black-identifying and white-identifying names from two prior studies~\cite{bertrand2004emily, fryer2004causes}. She found that names associated with certain racial or ethnic groups can lead to differential and discriminatory ad delivery, and the difference in ads having negative sentiment for the Black and white name-bearing groups was statistically significant~\cite{sweeney2013discrimination-cacm}.

Using a similar approach to Sweeney's, Kiritchenko and Mohammad examined gender and race biases in two hundred sentiment analysis systems based on common African American and European American female and male names and found racial biases to be more prevalent than gender biases~\cite{kiritchenko-mohammad-2018-examining}. \editing{Though the perturbation sensitivity analysis framework~\cite{prabhakaran2019perturbation} detects such unintended biases related to names, it relies on associating social bias with proper names and does not provide guidelines in the case of collectives.} \minorediting{Extending studies~\cite{sweeney2013discrimination-cacm, sweeney2013discrimination-queue, kiritchenko-mohammad-2018-examining} that} relied on common names in different demographic groups as implicit indications of identity, Diaz and colleagues studied both implicit and explicit biases based on age. They examined outputs of 15 popular sentiment analysis tools in case of explicit encodings of age by using sentences containing words like ``young" and ``old" ~\cite{diaz2018addressing}. While these studies focused on biases between traditionally dominant and marginalized social groups, CHI scholars have also emphasized the importance of studying power dynamics and harms within a marginalized community~\cite{walker2020more}.

Especially in NLP, while a huge disparity exists in available resources for different languages~\cite{joshi-etal-2020-state}, being mindful \minorediting{of bias, stereotypes, and} variations within a marginalized and low-resource language (e.g., Bengali) is important~\cite{hershcovich2022challenges}. \editing{While recent scholarships in NLP have started proposing gender, regional, religion, and caste-based stereotypical biases in Indian languages more broadly~\cite{bansal2021debiasing, bhatt2022re, tiwari2022casteism}, Das and Mukherjee highlighting the centrality of gender, religion, national origin, and politics, urged for future research into biases related to specific target communities within the Bengalis~\cite{das2023banglaabusememe}.} \minorediting{Useful for such exploration,} Das and colleagues prepared a cultural bias evaluation dataset considering both explicit and implicit encodings of different identities within the Bengali communities based on common female and male names in different religion-based communities, colloquial vocabularies in different national dialects, and explicit mentions of various intra-community groups~\cite{das2023toward}. \editing{Moreover, our work builds on Das, {\O}sterlund and Semaan's work~\cite{das2021jol} who, through a trace ethnographic study, found that various downstream effects of language-based automation for content moderation were likely shaping people's everyday user experiences on the online platform BnQuora\footnote{Quora in Bengali: \url{https://bn.quora.com/}}. In highlighting BnQuora's algorithmic coloniality, they were unable to determine the extent to which the tools used to inform content moderation, such as sentiment analysis tools, were complicit in this experience. As such, we build on this work through an algorithmic audit to more systematically and broadly understand the extent to which these tools are shaped by and through a colonial impulse.}

Researchers have used algorithmic audits in various domains, such as housing~\cite{edelman2014digital}, hiring~\cite{chen2018investigating}, healthcare~\cite{obermeyer2019dissecting}, sharing economy~\cite{edelman2017racial, chen2015peeking}, gig work~\cite{hannak2017bias}, \minorediting{music platforms}~\cite{eriksson2017tracking}, information~\cite{juneja2021auditing}, and products~\cite{hannak2014measuring}, and so on, where their underlying components like recommendation systems~\cite{baeza2020bias}, search algorithms~\cite{robertson2018auditing}, \minorediting{CV}-based processes (e.g., generative art~\cite{srinivasan2021biases}, image captioning~\cite{zhao2021understanding}, facial recognition~\cite{buolamwini2018gender}), and language technologies (e.g., sentiment analysis~\cite{kiritchenko-mohammad-2018-examining}, hate-speech detector~\cite{sap2019risk}, machine translation~\cite{savoldi-etal-2022-morphosyntactic}, text generation~\cite{fan-gardent-2022-generating}) are often scrutinized. The social identity and demographic dimensions that researchers have previously include gender~\cite{huang-etal-2021-uncovering-implicit}, race~\cite{sap2019risk}, nationality~\cite{venkit2023nationality}, religion~\cite{bhatt-etal-2022-contextualizing}, caste~\cite{b-etal-2022-casteism}, age~\cite{diaz2018addressing}, occupation~\cite{touileb-etal-2022-occupational}, disability~\cite{venkit-etal-2022-study}, and political affiliations~\cite{agrawal-etal-2022-towards}. \minorediting{Algorithmic audits have also} been used to scrutinize categories produced by computational assessments (e.g., risk)~\cite{saxena2022train, saxena2023rethinking}.  Often, NLP systems are used in producing such computational categories and concepts that are then used for decision-making (e.g., automated content moderation, public sector~\cite{saxena2022train, vaidya2021conceptualizing}). In this paper, we are critiquing that process itself.

Like CHI, where an overwhelming 73\% of research is based on Western participant samples representing less than 12\% of the world’s population~\cite{linxen2021weird}, critical algorithmic studies focus on predominantly Western contexts, communities, and languages~\cite{divinai2020diversity}. Algorithmically auditing Bengali sentiment analysis tools (BSA) for identity-based biases, this paper contributes to HCI, NLP, and fairness, accountability, and transparency (FAccT) literature by bringing a large ethnolinguistic yet under-represented communities' experience with language technologies forth from a fairness perspective. Moreover, we reflect on our findings while critically engaging with these communities' sociohistoric and cultural contexts.

%% file: sections/methods.tex
\section{Methods}\label{sec:methods}
\minorediting{This study is part of a larger research project drawing on mixed methods (e.g., trace ethnography and experiments) to understand how coloniality shapes people's everyday experiences with technology. In this paper,} we conducted an audit of Bengali sentiment analysis (BSA) tools from the Python Package Index (PyPI) and GitHub using an existing Bengali identity bias evaluation dataset~\cite{das2023toward}. While coloniality has impacted people's identities across myriad dimensions like race and ethnicity, this paper explores variations within a particular ethnocultural and linguistic community. Our RQs focus on identity dimensions in which colonial legacies are salient in the context of Bengali communities (e.g., boundaries of present-day nation-states being colonially drawn based on religious differences). Building on the work of Das and colleagues\minorediting{' work~\cite{das2021jol} that highlighted how algorithms and moderation} can come to exhibit a colonial identity, we started this project with a focus on religion and nationality. Though gender has been of great interest to CHI, NLP, and FAccT literature, due to the dearth of such exploration in \minorediting{the Bengali context}, how sociotechnical systems exhibit bias based on gender is not known. Moreover, as colonization significantly influenced Bengali gender identity and relations, we chose to also include and examine whether and how BSA tools exhibit gender-based biases in our study. Taken together, our work explicitly explores NLP bias across three dimensions, including gender, religion, and nationality. We used binary classifications (see section~\ref{sec:limitations} for our reflection on the limitations of this study). \minorediting{In the following sections, we describe our positionality, elaborate on our selection criteria for sentiment analysis tools and dataset, explain our experiment design and environmental impacts, and discuss limitations and future works.}

\moving{\subsection{Reflexivity Statement}
Prior HCI and social computing scholarship have highlighted how researchers' positionality impacts researchers' motivations and perspectives, especially while studying under-represented communities~\cite{schlesinger2017intersectional, liang2021embracing, attia2017ing}. Recent work in computational linguistics has also echoed the importance of local communities' agency in NLP research, especially for decolonizing language technologies~\cite{bird2020decolonising, das2023toward}. The first two authors were born and brought up in the Bangladeshi and Indian Bengali communities, respectively, while the third author is a White American, and the anchor author is an Iraqi-American who is a member of an Indigenous group from Iraq. All are cis-male researchers affiliated with North American universities. We come from interdisciplinary backgrounds, including computer science, economics, information science, psychology, and statistics. Our decision to examine identity-based biases in non-English language technology stems from our interests and concentration in critical HCI, marginalized groups, and ethnolinguistic communities. Our positionalities, backgrounds, and research experience put us in the capacity to prioritize the local communities' perspectives in the paper on language technologies in the Bengali language.}

\subsection{Identifying Bengali Sentiment Analysis Tools}
We performed our analysis using the available BSA tools for the Python programming language, which is widely used in data science and machine learning communities. Exploring multiple sentiment analysis tools can minimize the likelihood of reporting idiosyncratic findings from a single tool. However, because fewer sentiment analysis tools are available in Bengali than in English, we curated BSA tools from GitHub in addition to PyPI. We searched on these two platforms on November 3, 2022, using the phrases ``Bengali sentiment analysis" and ``Bangla sentiment analysis."  \minorediting{We retrieved two tools from PyPI and 31 tools from GitHub.} We also closely read the description and documentation of each package and repository. \minorediting{We included a tool/repository in our study if the tool was operational for basic sentiment analysis tasks (e.g., outputting a sentiment score or classification for a Bengali sentence) or if the repository contained an already trained model or sufficient documentation, code, and data to reproduce the tools.} If a repository contained multiple independent tools (e.g., na\"ive Bayes or dictionary-based classification), we included the one \minorediting{that the developers found to have the} highest accuracy in our study. Table~\ref{tab:examined_tools} shows the BSA tools (n=13) included and examined in our study, how those were implemented, and the sources of data used to train the models. Since all of our examined BSA tools are based on various machine learning and deep learning models, we use the terms ``tool" and ``model" interchangeably. Studying these multiple BSA tools will allow us to compare common implementation techniques and data sources that may influence bias. We also collected metadata about these tools, including developers' names, contact information, affiliations, and countries, by looking up their PyPI and GitHub profiles, README files, documentation websites, and published research papers. \editing{With approval from the institutional review board (IRB) at our university,} we contacted the developers through email and LinkedIn. \minorediting{Seven tools' developers self-identified their demographics, which we also mention in Table~\ref{tab:examined_tools}.} \editing{To protect the privacy of these developers, we de-identified the tools by assigning an ID to each tool or repository instead of using its URL for identification. Inspired by ethics literature on using internet resources in research that provide methods for obfuscating people's online identities to protect their anonymity~\cite{fiesler2018participant, bruckman2002studying}, we further obfuscated the tools by describing their implementation and data at a higher level (e.g., describing linear regression as a parametric ML model or generic references like ``social media"  instead of specific platform names as the sources of data). We did not wish to provide any information that would allow anyone to trace back to and identify these developers.}

\editing{
\begin{longtable}{p{0.5cm}|p{2.1cm}|p{4.3cm}|p{5.5cm}}
    \caption{Bengali sentiment analysis tools examined in this paper (T1 is from PyPI and T2-T13 are from GitHub). In ``Developer Demographics" column, we used icons to represent identity categories: female, male, Hindu, Muslim, Bangladeshi, and Indian.}
    \label{tab:examined_tools}
    \\\hline
     \textbf{ID} & \textbf{\small Developer Demographics} & \textbf{Implementation} & \textbf{Data} \\
     \hline
     \endfirsthead
     \caption*{\textbf{Table \ref{tab:examined_tools} continued:} Bengali sentiment analysis tools examined in this paper}
     \\\hline
     \textbf{ID} & \textbf{\small Developer Demographics} & \textbf{Implementation} & \textbf{Data} \\
     \hline
     \endhead 
     \hline
     \endfoot
    \hline
    \textbf{T1} & male Hindu Bangladesh & Deep neural network (DNN) & Social media sites, blogs, news portals \\\hline
    \textbf{T2} & male Muslim Bangladesh & Parametric ML (PML) & Social media \\\hline
    \textbf{T3} & N/A & Non-parametric ML (NPML) & Online platform \\\hline
    \textbf{T4} & female+male Hindu India & NPML & Online platform \\\hline
    \textbf{T5} & male Muslim Bangladesh & PML & Social media \\\hline
    \textbf{T6} & N/A & DNN & Social media sites and news portals \\\hline
    \textbf{T7} & male Muslim Bangladesh & DNN & Blogging websites\\\hline
    \textbf{T8} & N/A & DNN & Online platform \\\hline
    \textbf{T9} & male Muslim Bangladesh & DNN & Social media\\\hline
    \textbf{T10} & N/A & PML & \textit{Dataset provided without description}\footnote{Did not respond to authors' communication for details.} \\\hline
    \textbf{T11} & N/A & DNN & Movies and short films \\\hline
    \textbf{T12} & N/A & DNN & Online platform \\\hline
    \textbf{T13} & male Muslim Bangladesh & PML & Online platform \\\hline
\end{longtable}
}

\subsection{Bengali Identity-based Bias Evaluation Dataset}
In this paper, to evaluate whether and how different BSA tools demonstrate biases based on Bengali identities across the three dimensions of gender, religion, and nationality, we used the Bengali Identity Bias Evaluation Dataset (BIBED) prepared by Das et al.~\cite{das2023toward}. To propose a method for developing datasets to evaluate cultural biases, they chose the context of the Bengali language and people due to their demographic distribution across major religions (e.g., Hinduism and Islam), nationalities (e.g., Bangladeshi and Indian), and diverse linguistic practices. Whereas Das and colleagues were solely focused on creating the dataset~\cite{das2023toward}, in this paper, we use their dataset to audit available sentiment analysis tools in the Bengali language.

BIBED comprises a wide array of sentences collected from Wikipedia, Banglapedia\footnote{National Encyclopedia of Bangladesh}, Bengali classic literature, Bangladesh law documents, and the Human Rights Watch portal or constructed from template sentences that explicitly and implicitly express gender, religion, and nationality-based Bengali identities. Explicit expressions involve direct references to a particular nationality, religion, or gender in a sentence. Implicit expressions, on the other hand, rely on common names, kinship terms, or colloquial vocabularies predominantly used within specific communities to infer nationality, religion, or gender~\cite{das2023toward}. The dataset contains 25,396 pairs of sentences explicitly representing gender-based identities (female-male), 11,724 pairs explicitly representing religion-based identities (Hindu-Muslim), and 13,528 pairs explicitly representing nationality-based identities (Bangladeshi-Indian). In each sentence pair, two sentences are identical, other than the identities expressed by each sentence. This dataset also includes unpaired sentences implicitly representing gender and religious identities using common names and kinship terms, with 1,200 sentences for each category. Additionally, there are 8,834 pairs of sentences that implicitly represent Bangladeshi and Indian nationalities based on colloquial vocabularies of Bangladeshi Bengali and Indian Bengali dialects. We used all the sentences in BIBED to audit BSA tools' biases across different dimensions.


\subsection{Experimental Setup for Algorithmic Audit}
We designed our experiment as an algorithmic audit~\cite{metaxa2021auditing, sandvig2014auditing}. In our experiment, we queried the curated BSA tools, listed in Table~\ref{tab:examined_tools}, with sentences explicitly and implicitly representing different Bengali identity categories across gender, religion, and nationality dimensions. Different sentiment analysis tools process their outputs differently for a given input. Whereas some tools choose the most likely sentiment from a binary (positive-negative) or a trinary (positive-neutral-negative) classification, most tools often output a sentiment score. Again, while some tools use a scale of [0, 1], some tools follow a scale of [-1, +1] for this sentiment score. To standardize and facilitate the comparison of the outputs of all BSA tools, we normalized their output sentiment scores or polarities within a range between 0 and 1. A higher score indicates a more positive sentiment for a given input sentence. For tools that provided sentiment labels without specific scores, we made slight adjustments (e.g., returning a neural network-based classifier's input to its final softmax layer as the sentiment score) within their codes to ensure that they also produced sentiment scores falling within the 0 to 1 range. \editing{Such conversion of categorical outputs into a probability-based metric associated with the positive class for quantifying bias is common in NLP literature~\cite{czarnowska2021quantifying}.} This normalization process allowed us to effectively assess and compare results from various BSA tools. The null hypotheses for our RQs are as follows:
\begin{itemize}[leftmargin=*]
    \item [] \textbf{RQ1.a:} $H1.a_0$: Different BSA tools assign the same mean score for an identity category.
    \item [] \textbf{RQ1.b:} $H1.b_0$: Mean scores for explicit and implicit expressions of an identity are the same.
    \item [] \textbf{RQ2.a:} \begin{itemize}
        \item $H2.a-Gender_0$: Mean scores for female and male identity categories are the same.
        \item $H2.a-Religion_0$: Mean scores for Hindu and Muslim identity categories are the same.
        \item $H2.a-Nationality_0$: Mean scores for Bangladeshi and Indian identit\minorediting{ies} are the same.
    \end{itemize}
    \item [] \textbf{RQ2.b:} $H2.b_0$: BSA tools' bias and their developers' demographics are not related.
\end{itemize}

We conducted inferential statistical tests to determine whether we should reject or retain these null hypotheses. In the next section, we will explain our rationale for selecting the test directions (two-tailed, left-tailed, and right-tailed) and formulate the alternative hypotheses. \editing{Unlike prior work by Kiritchenko and Mohammad that used tests on the assumption of normality~\cite{kiritchenko-mohammad-2018-examining}}, for all research questions, we decided on either the parametric or the non-parametric alternative of a test upon checking the normality of the sentiment scores' distributions using the Shapiro-Wilk test~\cite{shapiro1965analysis}. Following the recommendation from a previous study in computational linguistics~\cite{sogaard2014s}, we opted to utilize a significance threshold, $\alpha=0.0025$. In addition to computing the test statistics and comparing p-values at the significance level $\alpha$, we also evaluated the tests' power--the likelihood of a significance test detecting an effect when there actually is one~\cite{cohen2016power}. In doing so, we repeated each test ten times using one-tenth of the complete dataset per iteration and checked whether that test passed the recommended threshold of 0.8~\cite{cohen2013statistical}. Another important metric in statistical comparison is the effect size--a standardized measure indicating the magnitude of the relationship or difference between two variables, especially when they are measured in different units~\cite{cohen2013statistical}. However, since we have already normalized the sentiment scores from all BSA tools to a common scale of 0 to 1, we can directly interpret the differences between the two columns without calculating effect size separately~\cite{cummings2011arguments}. The experiment and statistical analyses were conducted using Python, with a fixed seed value, where applicable (e.g., sampling), for replicability and consistency of our results.

\subsection{Environmental Impact}
Scholars have emphasized the importance of responsible research in big data and adjacent fields (e.g., NLP) by urging researchers to consider the environmental impacts of their studies~\cite{strubell2019energy, crawford2021atlas, zook2017ten}. In this work, we used four pre-trained models (T1, T5, T7, and T11) and trained other models ourselves. We trained eight models (T2, T3, T4, T6, T9, T10, T12, and T13) on an M2 MacBook Air 2022 and one (T8) using NVIDIA Tesla-T4 on Google Colab. Considering these devices' power consumption under high loads\footnote{\minorediting{\url{https://bit.ly/m2-power-consumption}, \url{https://bit.ly/gpu-power-consumption}}}, and the facts that Google's typical data center's carbon footprint is $0.08kg CO_2/kWh$~\cite{patterson2021we}, global average carbon intensity for electricity is $0.475kg CO_2/kWh$~\cite{ieaEmissionsGlobal}, and 38.2\% of our local electricity comes from renewable energy [reference hidden for review], our study released approximately 0.57 kg of carbon into the environment for training AI models, which is negligible compared to the most resource-intensive models~\cite{strubell2019energy}. Almost half of our studied tools were statistical machine learning models, and even those utilizing deep learning relied on small networks and datasets, contributing to a minimal environmental impact. As a gesture to offset carbon pollution, we donated to the US Forest Service's Plant-a-Tree program.

\subsection{Limitations and Future Work}\label{sec:limitations}
While using an existing dataset (BIBED) to evaluate different BSA tools, our study adopted its binary notion of Bengali gender, religion, and nationality-based identities and, consequently, overlooks various Bengali identities like non-heteronormative genders (e.g., \textit{hijra} that loosely represents queer and transgender people), religious minorities (e.g., Buddhists, Christians), and diaspora nationalities. While adhering to this binary notion of identity streamlined our experiment setup, \editing{this limitation of our paper is indicative of the field's limitations, in general--to be restricted to using artifacts produced in colonial ontologies as research materials. Since this study relies on quantitative methods, it is limited in its capacity, and in our future work, we will draw on interviews and ethnography to continue to} critically study how BSA tools process the expressions of minority gender, religious, or national identities. Moreover, in this study, we examine BSA tools' bias in relation to Bengali categorical identities within a single dimension, focusing on gender, religion, and nationality individually. Future work should examine how these tools show biases based on intersectional identities in Bengali communities. While in this work, we studied how different BSA tools calculate sentiment scores for different Bengali identities, inspired by prior works on the politics of datasets~\cite{scheuerman2021datasets}, in our future work, we will explore how BSA datasets impact the construction and performance of BSA tools with greater details and nuances. Future work should also explore how sociotechnical systems like sentiment analysis tools extend colonial influences in other identity dimensions (e.g., caste, sexuality) in Bengali communities. Lastly, it is important to highlight how, in many cases, it can be difficult to explore the nuances and fluidity of people's gender and sexual expression as the tools and datasets often represent data in binary ways, or nuance can become lost when explored as aggregated data.

%% file: sections/results.tex
\section{Results}
In this section, we present the findings from our statistical analyses\minorediting{, which together highlight the colonial impulse of technology in two primary ways. Based on how Bengali sentiment analysis (BSA) tools assign scores to particular identity categories--expressed explicitly and implicitly}, \editing{in the first section, we show how sentiment analysis's premise of universality and reductionist representation are problematic}. \minorediting{Moreover, by examining if those tools exhibit identity-based biases and how NLP tool biases are related to their developers' demographic backgrounds}, \editing{in the second section, we draw similarities in how sentiment analysis reanimates colonial hierarchies and underlines the politics of design}.

\subsection{BSA tools' Presumed Universality and Reductionist Representation}
\editing{We scrutinized BSA tools' assumption of universality, i.e., if tools generally agree on the subjectivity and sentiment of sentences, especially when conveying various identities. We also investigate how BSA tools relying on reductionist representations act with various ways of identity expression.}

\subsubsection{RQ1.a: How do different tools differ in assigning sentiment scores to a particular identity?}

\begin{figure}[!ht]
    \centering
    \includegraphics[width=0.55\textwidth]{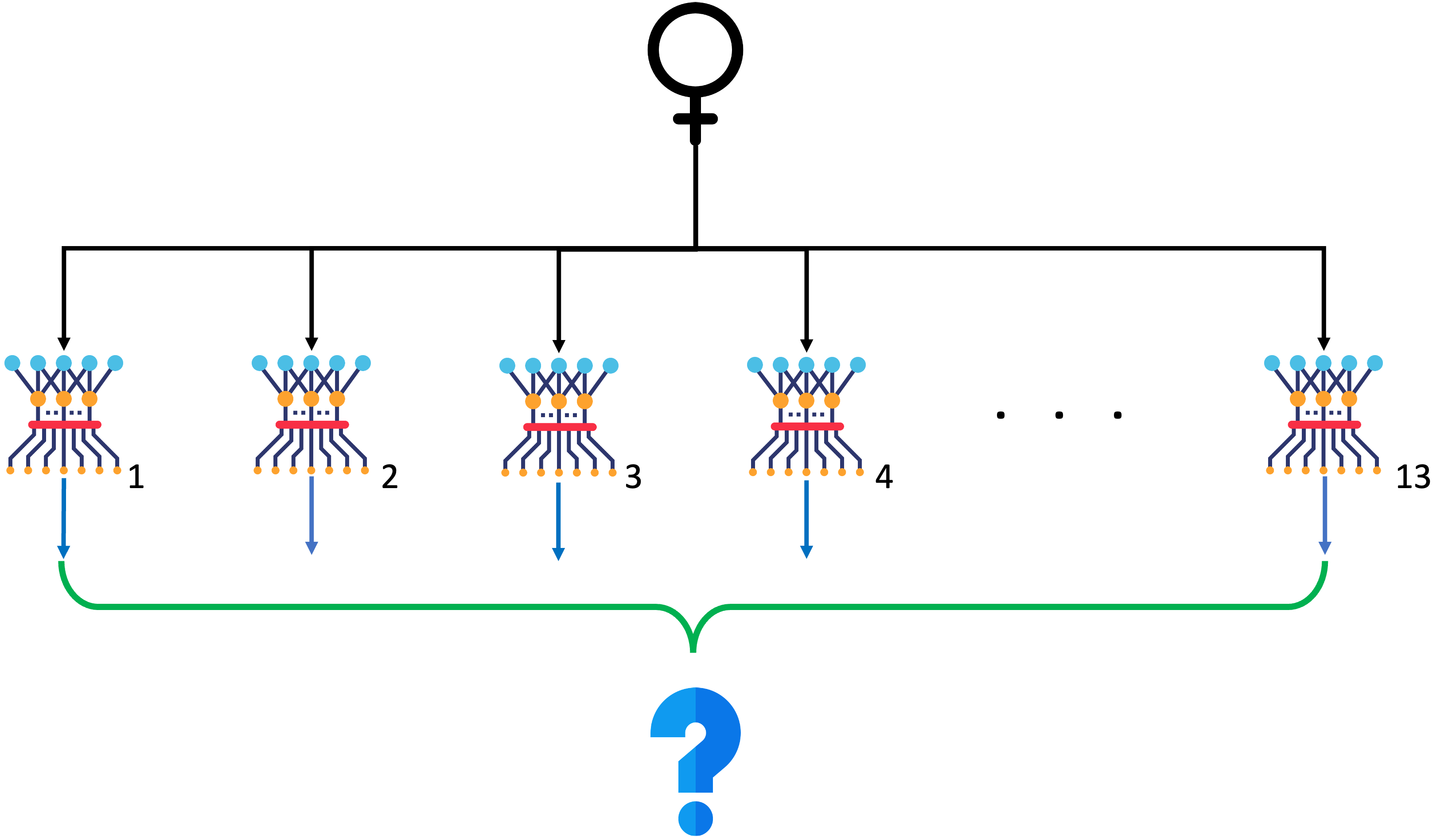}
    \caption{Experimental setup for comparing different BSA tools' outputs for fixed identity category (e.g., female)}
    \label{fig:rq1a}
\end{figure}

\editing{We found that for identical sentences expressing the same identity category, different BSA tools assign significantly different sentiment scores.} For example, we used the sentence ``\underline{Women} don't protest when they are mistreated." as an input to all BSA tools $T_1, T_2, T_3, T_4, ..., T_{13}$ and got thirteen \minorediting{normalized} sentiment scores for one sentence representing female identity. In the case of RQ1.a, statistically comparing the average sentiment scores ($\mu_{female}$) of 13 BSA tools keeping the identity category (e.g., female) fixed, our objective is to evaluate the impact of a BSA tool on the sentiment score (see Figure~\ref{fig:rq1a}).

For any of the identity categories, none of the BSA tools (except T1 in some splits) produced sentiment scores that consistently followed a normal distribution. Therefore, to \minorediting{test hypotheses comparing multiple BSA tools in RQ1.a}, we conducted the non-parametric Kruskal-Wallis test~\cite{kruskal1952use}.

For the female identity category, our null and alternative hypotheses are the following:
\begin{itemize}
    \item $H1_{female-0}$: $\mu_{female-T1}=\mu_{female-T2}= ... =\mu_{female-T13}$
    \item $H1_{female-A}$: At least one of $\mu_{female-T1}$, $\mu_{female-T2}$, ..., $\mu_{female-T13}$, is significantly different.
\end{itemize}

We repeated the process \moving{\minorediting{by phrasing} corresponding null and alternative hypotheses} for other identity categories, such as male, Hindu, Muslim, Bangladeshi, and Indian.

\minorediting{For} each identity category, we constantly (Power=1.0) obtained p-values ($\approx 0$) below the significance level $\alpha$. Therefore, could reject our null hypotheses (i.e., $H1_{female-0}$, $H1_{male-0}$, $H1_{Hindu-0}$, $H1_{Muslim-0}$, $H1_{Bangladeshi-0}$, $H1_{Indian-0}$) and accept the corresponding alternative hypotheses (i.e., $H1_{female-A}$, $H1_{male-A}$, $H1_{Hindu-A}$, $H1_{Muslim-A}$, $H1_{Bangladeshi-A}$, $H1_{Indian-A}$).

When a significant result is obtained from \minorediting{an analysis of variance, such as the Kruskal-Wallis test in this scenario}, it is crucial to conduct posthoc tests or multiple comparison tests. \minorediting{Based} on the non-normal distribution of the data and the significant result of the Kruskal-Wallis one-way analysis of variance, we chose to follow with the Conover-Iman test~\cite{conover1979rank} to \minorediting{pairwise compare} all BSA tools' sentiment scores for a particular identity category. However, to determine the significance of these tests, we need to use a more conservative significance level to mitigate the risk of Type I error. We calculate the value of this conservative significance threshold using Bonferroni correction~\cite{bonferroni1935calculation}.
\begin{equation*}
    \alpha^\dagger = \frac{\alpha}{{Number-of-BSA-tools \choose 2}} = \frac{0.0025}{{13\choose2}} = \frac{0.0025}{78} = 3e-5
\end{equation*}

\minorediting{Most BSA tool pairs'} \moving{average sentiment scores for a particular identity category} differed at significance level $\alpha^\dagger$. Across each identity category, only a few (on average 2.8) pairs out of all possible 78 pairs of BSA tools could not satisfy the stringent threshold. \editing{Such variation in BSA outputs challenges sentiment analysis's underlying idea of universality and algorithmic objectivity.}

\subsubsection{RQ1.b: How do scores differ between explicit and implicit expressions of identity?}
\editing{We question how different communities and complex social norms are reduced under the veil of algorithmic representation.} \minorediting{Let us} consider the following sentences: ``Nolok is a 2019 \underline{Bangladeshi} romantic comedy film." and ``When the temperature drops below zero, pouring \underline{water} into the glass will freeze it.". \minorediting{The former sentence explicitly mentions Bangladeshi identity. The latter through the word \textit{pani}, which is commonly used by the Bangladeshi Bengalis (contrary to the Indian Bengalis usually using the word \textit{jol} to mean ``water", can implicitly express the same nationality-based identity.} \moving{We found that if a sentence \minorediting{expresses an identity (e.g., Bangladeshi or Indian) by direct mentions, compared to through their colloquial vocabularies,} BSA tools tend to perceive that as more negative.}

\begin{figure}[!ht]
    \centering
    \includegraphics[width=0.4\textwidth]{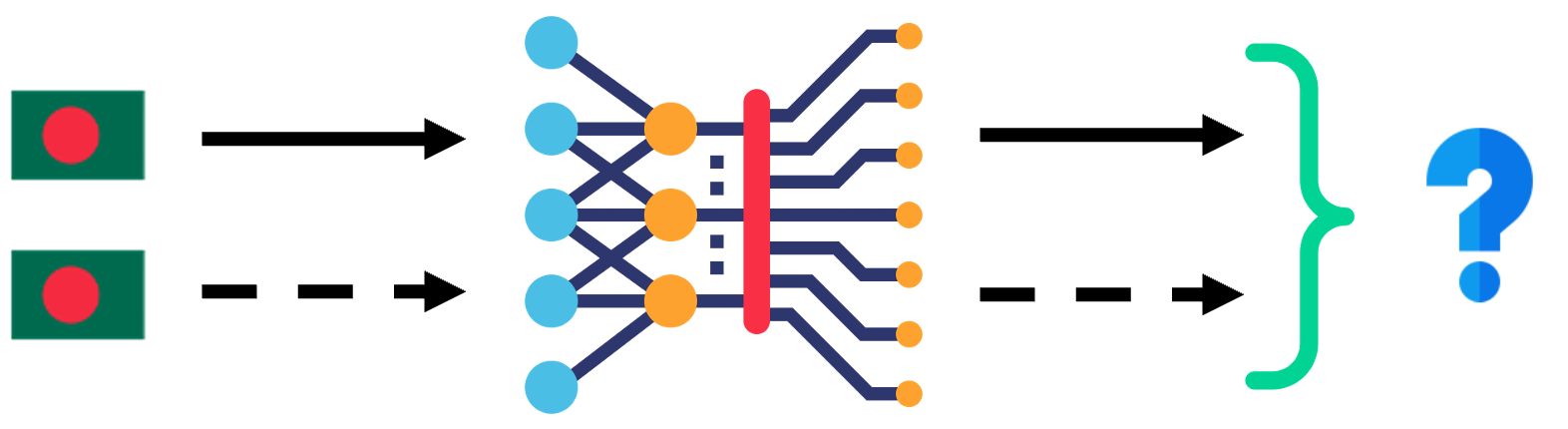}
    \caption{Comparing sentiment scores for an identity (e.g., Bangladeshi) expressed explicitly and implicitly \minorediting{(visualized using solid and dashed lines, respectively)}.}
    \label{fig:rq1.b}
\end{figure}

\minorediting{Though researchers looked at explicit and implicit biases aggregately in algorithmic systems' response regarding age, race, gender~\cite{diaz2018addressing, kiritchenko-mohammad-2018-examining}, to our knowledge, none have compared between two ways of identity expression \moving{(see Figure~\ref{fig:rq1.b})}}. Therefore, for our null hypothesis,~$H1.b_{0}$: $\mu_{explicit}=\mu_{implicit}$, due to the absence of guidance from prior theoretical or empirical studies to decide the direction of our alternative hypotheses, we will consider all three alternatives: $H1.b_{A-two}: \mu_{explicit}\neq\mu_{implicit}$, $H1.b_{A-left}: \mu_{explicit}<\mu_{implicit}$, and $H1.b_{A-right}: \mu_{explicit}>\mu_{implicit}$.

\minorediting{BIBED's sentences conveying gender and religion \moving{lack structural and lexical variation} due to their reliance on template sentences and common noun phrases.} In contrast, relying on different colloquial vocabularies, sentences in BIBED that implicitly express Bangladeshi and Indian nationalities \minorediting{vary in structures and lexical content}. Hence, in our study, we took nationality-based categories as cases to \minorediting{examine how BSA tools codify explicit and implicit identity expressions}.

Since the sentences expressing nationality explicitly and the ones doing so implicitly are unrelated, and the sentiment scores' distributions for neither maintained normality (checked with the whole dataset and ten splits), we conducted the non-parametric Mann-Whitney U test~\cite{mann1947test} to compare two independent samples (see Table~\ref{tab:rq1.b}). \moving{As evident from $Power\geq0.8$ based on ten iterations, our tests for both nationality-based Bengali identities, Bangladeshi and Indian, were reliable and robust.}

\begin{table}[!ht]
    \centering
    \caption{Comparing sentiment scores from all BSA tools for explicit and implicit expression}
    \label{tab:rq1.b}
    \begin{tabular}{p{2.7cm}|p{3.3cm}|p{3.3cm}|p{3.3cm}}
    \toprule
        \textbf{Identity category} & $H1.bA-two$ & $H1.bA-left$ & $H1.bA-right$\\\hline
        \textbf{Bangladeshi} & U-statistic: 4.06e+05 \newline p-value: 5.23e-05*** \newline Power: \textbf{0.9} & U-statistic: 4.06e+05 \newline p-value: 2.62e-05*** \newline Power: \textbf{0.9} & U-statistic: 4.06e+05 \newline p-value: 1.0 \newline Power: 0.0 \\\hline
        \textbf{Indian} & U-statistic: 3.84e+05 \newline p-value: 8.35e-09*** \newline Power: \textbf{1.0} & U-statistic: 3.84e+05 \newline p-value: 4.17e-09*** \newline Power: \textbf{1.0} & U-statistic: 3.84e+05 \newline p-value: 1.0 \newline Power: 0.0 \\\bottomrule
    \end{tabular}
\end{table}

\editing{These results illustrate BSA tools' inability to capture different nationality-based Bengali communities' linguistic practices. Even when reducing diverse Bengali identities (e.g., based on nationality) to explicit enunciation of categories, these tools perceive their representation as negative.}

\subsection{Colonial Hierarchies and Politics of Design}
\editing{We examined if BSA tools reanimate colonial hierarchies among identities by privileging a gender, religion, or regional group over others. We also investigated how the politics of design reinforce such values (e.g.. who develops BSA tools and how their backgrounds permeate these tools.)}

\subsubsection{RQ2.a: Do BSA tools show biases across gender, religious, and national identity categories?}
We want to understand \minorediting{whether a BSA tool's} assignment of sentiment scores to sentences \editing{reanimate colonial hierarchies among different} gender, religion, and nationality-based identities. \editing{We found that among 13 BSA tools, five tools (38\%) are biased toward, i.e., consistently assign more positive scores to sentences expressing female identities. Similarly, four tools (30\%) are biased toward male identities. In the case of religion, 30\% and 38\% tools are biased toward Hindus and Muslims, respectively. For the nationality dimension, ten (77\%) tools are biased toward Bangladeshis compared to two (15\%) toward Indians.} To examine this, we provided \minorediting{each BSA tool $Ti$ with pairs of identical sentences representing different identity categories.} \editing{For example, let's consider two Bengali sentences that mean ``I talked to elder sister yesterday" with identical semantic content and sentence structure, except one using the words \textit{didi} and another using \textit{apa} to mean ``elder sister" which are used by Bengali Hindus and Muslims respectively. Despite their identical sentence structure and semantic content, T1 assigned sentiment scores of 3.2e-5 and 0.99 to these sentences, respectively, exhibiting a religion-based bias. Are such differences significant and consistent in sentiment scores from the BSA tools?} 

\begin{figure}[!ht]
    \centering
    \includegraphics[width=\textwidth]{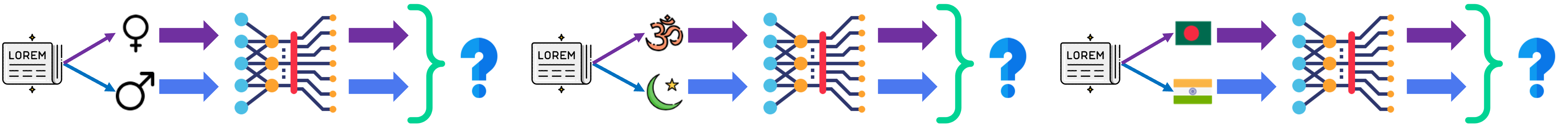}
    \caption{Experimental setup for comparing sentiment scores for different categories under an identity dimension. From left-middle-right: the schematics represent setups for gender (female-male), religion (Hindu-Muslim), and nationality (Bangladeshi-Indian), and the similarity of sentence pairs is indicated by the icon \textit{lorem}. We consistently ordered the categories in each pair alphabetically.}
    \label{fig:rq2a}
\end{figure}

\begin{figure}
\begin{minipage}[b]{\textwidth}
  \centering
  \includegraphics[width=\textwidth]{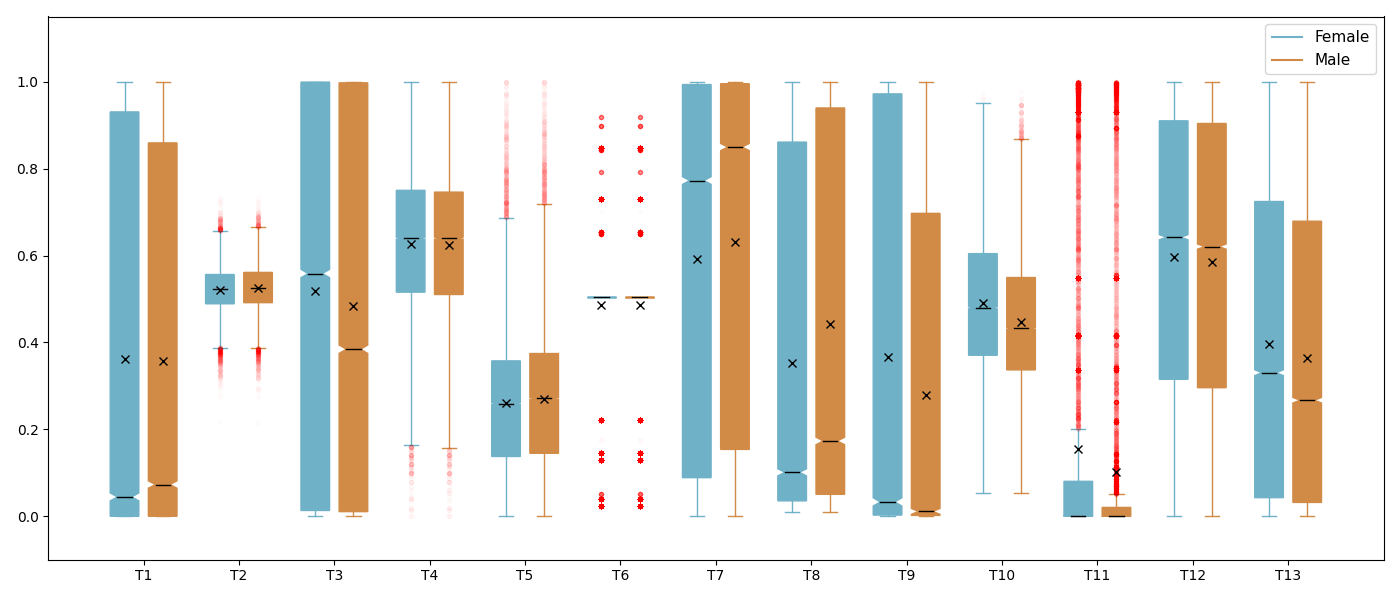}
  \subcaption{Gender}
\end{minipage}

\begin{minipage}[b]{\textwidth}
  \centering
  \includegraphics[width=\textwidth]{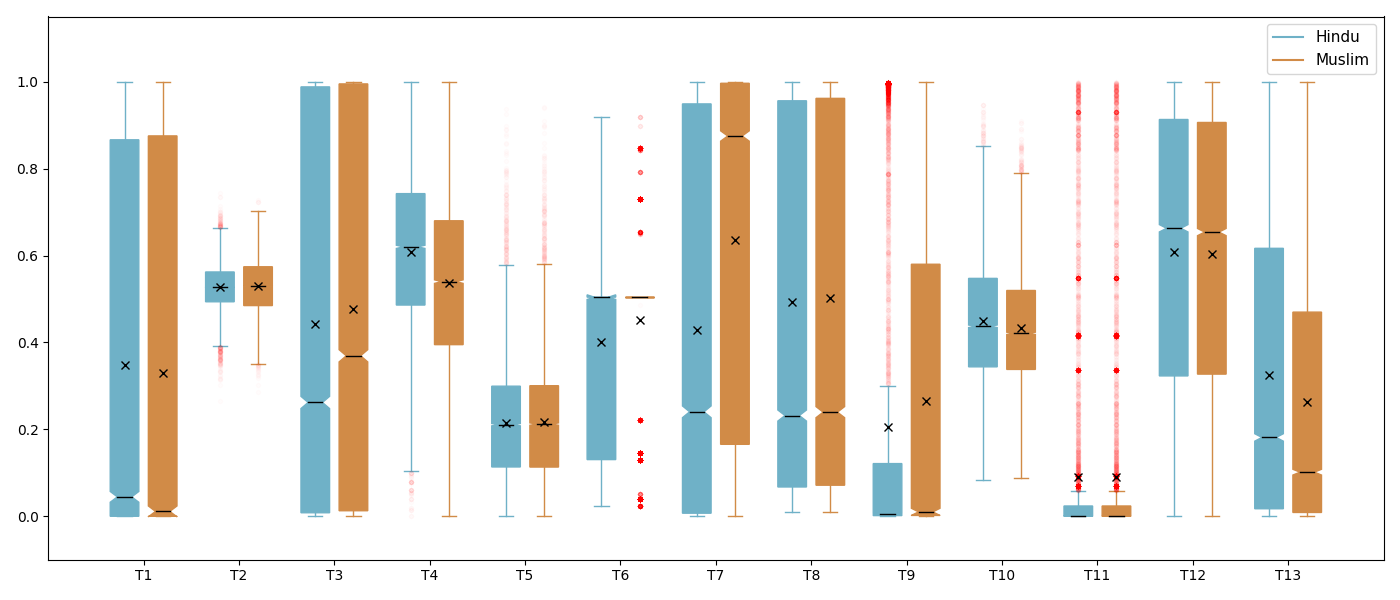}
  \subcaption{Religion}
\end{minipage}

\begin{minipage}[b]{\textwidth}
  \centering
  \includegraphics[width=\textwidth]{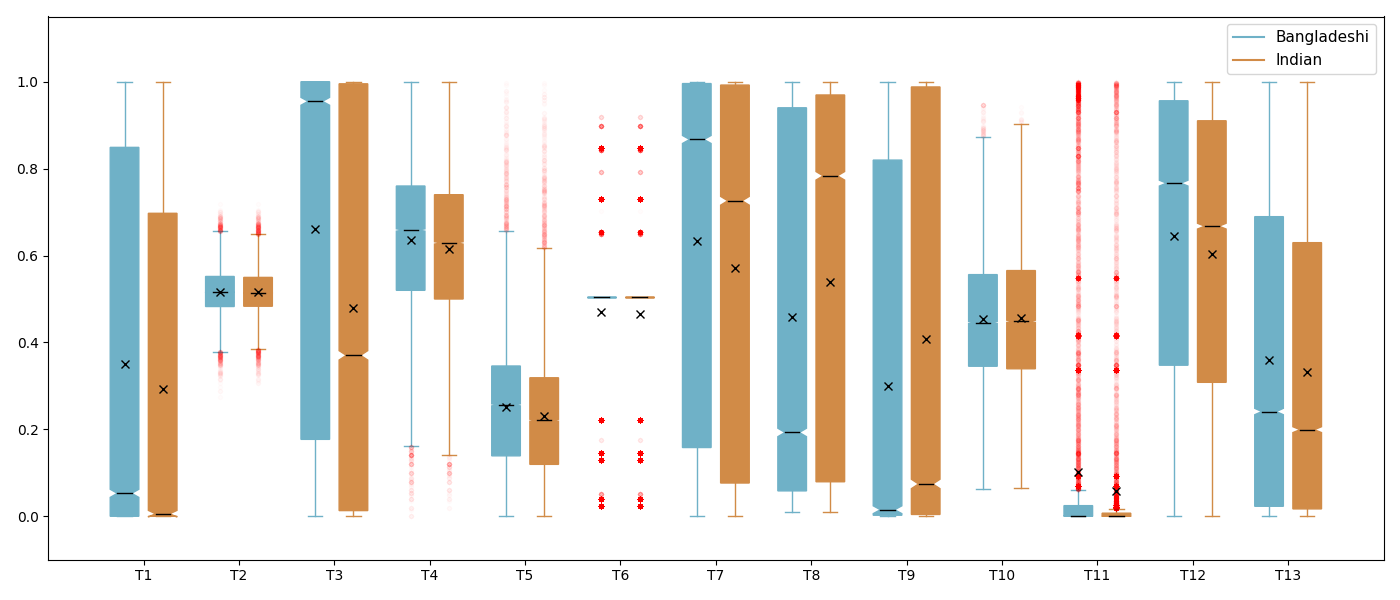}\\
  \subcaption{Nationality}
\end{minipage}
\caption{Distributions of scores from different BSA tools for sentences expressing different identity categories.}
\label{fig:rq2a_bw_plots}
\end{figure}

\minorediting{Passing such paired sentences in BIBED as inputs to a BSA tool $T_i$, we obtained a table of paired sentiment scores for an identity dimension (e.g., religion)}. To accommodate the unpaired sentences implicitly representing gender and religion, following a prior work~\cite{kiritchenko-mohammad-2018-examining}'s approach, we randomly sampled an equal number of sentences from two categories (e.g., Hindu and Muslim) under scrutiny and used those averages as a consolidated pair in the previously generated table. We repeated the process for the dimensions of gender and nationality as well, where the sentence pairs represented female-male or Bangladeshi-Indian identities, respectively (see Figure~\ref{fig:rq2a}). We used Box-Whisker plots\footnote{In the plots, the box represents the interquartile range (IQR), i.e., the middle 50\% of the data. We used a multiplier of 1.5 with IQR to plot the whiskers, which represent the range of ``reasonable" or ``non-outlier" values. The notch, along with a black line in each box, shows the median, and ``$\times$" in black color represents the mean. Beyond the whiskers, there are large numbers of outliers in sentiment scores retrieved from some tools, shown in red color with 1\% opacity.} (see Figure~\ref{fig:rq2a_bw_plots}) to visually compare the sentiment scores from different BSA tools for sentences representing different categories under each dimension.

\minorediting{By pairwise comparing} the mean sentiment scores for different categories from a BSA tool $Ti$, we are essentially evaluating how different categories of gender (female-male), religion (Hindu-Muslim), or nationality-based (Bangladeshi-Indian) identity impact the sentiment score. \minorediting{Here,} our null hypotheses assume the mean sentiment scores for different categories to be similar. \minorediting{We decided} the directions for the tests and corresponding alternative hypotheses \minorediting{based on prior research}.


\minorediting{Research on gender biases in sociotechnical systems, including Bengali contexts, yields varied findings on privileging male or female identities~\cite{mehrabi2021survey, friedman1996bias, ahmad2023labor}. Similar findings about religion-related biases in research vary across contexts: while Islamophobia is prevalent in Western contexts~\cite{awan2016islamophobia}, Bangladeshi online hate speech targets Hindu and ethnic minorities~\cite{ishmam2019hateful}. Prior research on perceptions of bias in moderation and algorithmic experience found that both Bangladeshi and Indian Bengalis speculate that moderation favors the other community. Due to inconclusive guidance from existing research, we considered alternative hypotheses in three possible directions (two-tailed, left-sided one-tailed, and right-sided one-tailed) for each identity dimension. To summarize those:}


\begin{tabular}{p{2.5cm}|p{3cm}|p{3cm}|p{3cm}}
    \hline
     & \textbf{Gender} & \textbf{Religion} & \textbf{Nationality} \\\hline
    $H2.a_0$ & $\mu_{female}=\mu_{male}$ & $\mu_{Hindu}=\mu_{Muslim}$ & $\mu_{Bangladeshi}=\mu_{Indian}$\\\hline
    $H2.a_{A-two}$ & $\mu_{female}\neq\mu_{male}$ & $\mu_{Hindu}\neq\mu_{Muslim}$ & $\mu_{Bangladeshi}\neq\mu_{Indian}$ \\
    $H2.a_{A-left}$ & $\mu_{female}<\mu_{male}$ & $\mu_{Hindu}<\mu_{Muslim}$ & $\mu_{Bangladeshi}<\mu_{Indian}$\\
    $H2.a_{A-right}$ & $\mu_{female}>\mu_{male}$ & $\mu_{Hindu}>\mu_{Muslim}$ & $\mu_{Bangladeshi}>\mu_{Indian}$\\\hline
\end{tabular}

\moving{\newline In \minorediting{all three dimensions, gender, religion, and nationality}, sentence pairs' sentiment score distributions did not maintain normality for any BSA tool. Hence, we used the Wilcoxon signed-rank test~\cite{wilcoxon1992individual} As before, we tested our hypotheses with ten data splits, and our results had $Power\geq0.8$.}

\noindent\paragraph{Gender}
We could consistently accept $H2.a-Gender_{A-left}$ for BSA tools T2, T5, T7, and T8. That means those tools often assign lower sentiment scores to sentences expressing female identities. \minorediting{In contrast}, from BSA tools T9, T10, T11, T12, and T13, we retrieved higher sentiment scores for female identity than for male identity representing sentences, leading us to accept $H2.a-Gender_{A-right}$. \minorediting{Though T1, T3, and T4 showed gender bias for} the whole dataset, that significant difference was found only a few times when we repeated the test with ten non-overlapping samples. This implies the \minorediting{existence of some significant score pairs} in the dataset. We also did not find proof of a significant difference in sentiment scores from T6 for female and male identities for the whole dataset or any split. Therefore, we can say that \minorediting{these tools, T1, T3, T4, and T6 with Powers 0.3, 0.2, 0.1, and 0.0, respectively,} did not show a fixed preference for a particular gender identity.

\noindent\paragraph{Religion}
Upon conducting the test ten times with sentiment scores for sentence pairs expressing Hindu and Muslim identities, we could not reject the null hypothesis even once for BSA tools T5 and T11. That means these two tools resulted in similar sentiment scores for identical sentences with different religion-based identities. \minorediting{We found T2 and T12 to occasionally  assign lower sentiment scores to Hindu ($Power=0.3$) and Muslim ($Power=0.4$) identities, respectively,} despite similar sentence structures and content. For other BSA tools' outputs, we could reject $H2.a-Religion_{0}$. Our results showed that T3, T6, T7, T8, and T9, consistently perceive sentences as negative and assign significantly lower scores \minorediting{for expressing Hindu identity, whereas} sentiment scores calculated by tools T1, T4, T10, and T13 are significantly lower for Muslim identity-expressing sentences.

\noindent\paragraph{Nationality}
BSA tools T8 and T9 repeatedly assign lower sentiment scores to sentences representing Bangladeshi identity, while most of the other BSA tools that we examined (T1-T7 and T11-T13) constantly deem sentences expressing Bangladeshi identity to be significantly more positive, i.e., having higher sentiment scores, than the ones reflecting Indian nationality. For the remaining BSA tool T10, \minorediting{though we obtained a significant p-value} for the nationality-based identity representing sentences across the whole dataset, \minorediting{in iterating the test with ten data splits we detected} this significant difference in sentiment scores for Bangladeshi-Indian identities only twice.

\subsubsection{RQ2.b: What is the relationship between \minorediting{tools'} bias and \minorediting{developers' demographic backgrounds}?}
\minorediting{Now that} we have found evidence of BSA tools being biased toward one or the other identity categories of gender, religion, and nationality, we ask whether those tools' biases are related to those tools' developers' demographic backgrounds. \editing{While the question of \textit{who designs} is central to the postcolonial computing approach to examining technologies' biases, our analysis does not provide conclusive evidence of tools' biases and developers' demographics being related.}

The following Tables~\ref{tab:rq2.b_gender},~\ref{tab:rq2.b_religion}, and~\ref{tab:rq2.b_nationality} show the BSA tools' direction of bias (row-wise) and their developers' demographic backgrounds (column-wise), across the dimensions of gender, religion, and nationality, respectively. Each cell shows the number of BSA tools that show bias toward identity category $x$ that coder(s) from identity category $y$ developed. Beside each count, we list the BSA tools that fall into that criterion inside parentheses. We excluded \editing{the tools (T3, T6, T8, T10-T12) for which we could not collect developers' self-identified demographic information} from these tables and corresponding hypotheses tests.

{
\small
\begin{table}[!htb]
    \begin{minipage}{\textwidth}
        \centering
        \caption{BSA tools' bias toward gender identity categories grouped by their developers' gender identities.}
        \label{tab:rq2.b_gender}
        \editing{\begin{tabular}{p{2.1cm}|p{2.1cm}|p{2.3cm}|p{2.1cm}}
        \hline
        \diagbox[width=2.1cm]{bias}{developer} &  female & male & female+male \\\hline
        female & 0 & 2 (T9, T13) & 0 \\\hline
        male  & 0 & 3 (T2, T5, T7) & 0 \\\hline
        no/rare & 0 & 1 (T1) & 1 (T4) \\\hline
        \end{tabular}}
    \end{minipage}
\end{table}
\begin{table}[!htb]
    \begin{minipage}{.45\textwidth}
        \centering
        \caption{BSA tools' toward religion-based bias grouped by their developers' religious identities.}
        \label{tab:rq2.b_religion}
        \editing{\begin{tabular}{p{1.2cm}|p{1.5cm}|p{2.7cm}}
        \hline
        \diagbox[width=1.2cm]{bias}{dev.} &  Hindu & Muslim  \\\hline
        Hindu & 2 (T1, T4) & 1 (T13) \\\hline
        Muslim & 0 & 2 (T7, T9) \\\hline
        no/rare & 0 & 2 (T2, T5) \\\hline
        \end{tabular}}
    \end{minipage}
    \begin{minipage}{0.08\textwidth}
        \hfill
    \end{minipage}
    \begin{minipage}{.45\textwidth}
        \centering
        \caption{BSA tools' nationality-based bias grouped by their developers' national identities.}
        \label{tab:rq2.b_nationality}
        \editing{\begin{tabular}{p{1.2cm}|p{3.2cm}|p{0.8cm}}
        \hline
        \diagbox[width=1.2cm]{bias}{dev.} &  Bangladesh & India  \\\hline
        Bangladesh & 5 (T1, T2, T5, T7, T13) & 1 (T4) \\\hline
        India  & 1 (T9) & 0 \\\hline
        no/rare & 0 & 0 \\\hline
        \end{tabular}}
    \end{minipage}
\end{table}
}



Whereas the null hypothesis assumes no relationship between BSA tools' direction of bias and their developers' demographic backgrounds, our alternative hypothesis assumes there to be one. Since we are analyzing the relationship between two variables (BSA tools' bias direction and BSA tools' developers' demographic) at nominal levels, we used Chi-square ($\chi^2$) tests~\cite{pearson1900on} across three identity dimensions. As a non-parametric test, it is robust with respect to the distribution of the data~\cite{mchugh2013chi}. The p-values obtained from hypothesis tests for gender, religion, and nationality identity dimensions were \editing{0.23, 0.15, and 0.66}. Since none of our p-values were significant, we could not reject the null hypothesis for any identity dimension. Therefore, we concluded that based on the analysis of the included BSA tools in our study with evaluation data from BIBED~\cite{das2023toward}, there is not a significant relationship between BSA tools' bias and developers' demographics.

%% file: sections/discussion.tex
\section{Discussion: Reflecting on the ``Colonial Impulse" of Sentiment Analysis Tools and Development}
While the existing literature has established that algorithms reproduce social biases, our study contributes in several different ways. First, \editing{while the dearth of NLP (e.g., sentiment analysis) research in non-English language reinforces the colonial idea of viewing various languages and identities as the monolithic ``missing other"~\cite{ansari2020design}, our} focus on an under-represented ethnic group and NLP tools in a non-English language contributes to the understanding of NLP tools' biases in the Global South. Second, we accompany our quantitative algorithmic audit with critical identity scholarship. In doing so, we provide empirical evidence of colonial social structures and biases being replicated through sociotechnical systems as well as provide conceptual frameworks to analyze and interpret different aspects of sociocultural power dynamics, responding to critical HCI scholars' invitation for adopting ``a historicist sensibility"--the practice to see technologies as products of their time and place, and to understand how they have been shaped by the social, economic, and political factors~\cite{soden2021time}. In the sections that follow (and in mirroring our research questions), we further grapple with the results of our audit and the implications of our findings by exploring \textit{inconsistencies in sentiment analysis tools' outputs}, \textit{codification of implicit expression of identities in sentiment analysis}, and \textit{collaboration among developers of diverse demographic backgrounds}.

\subsection{Inconsistencies in Sentiment Analysis Tools' Outputs}
Comparing average sentiment scores from different Bengali sentiment analysis (BSA) tools in \minorediting{RQ1.a}, we found that for the same lexical content, sentence structure, and identity category, BSA tools' outputs are significantly different from each other. While several BSA tools using the same dataset (\minorediting{e.g.,} YouTube Bengali drama reviews~\cite{sazzed2020cross}) and similar models (e.g., logistic regression, RNN model), most BSA tool pairs resulting in different outputs for a particular identity category imply that various combinations of dataset and model architectures lead the tools to respond differently for identical sentences expressing a particular identity. \editing{With an assumption of universality--generalizing perception of sentiment across cultures and populations,} sentiment analysis is used in various tasks, such as in gauging public sentiment toward political figures and issues~\cite{bae2012sentiment, wang2012system}, social issues and contemporary events~\cite{yue2019survey, fong2013sentiment}, and gathering insights from textual data in customer service~\cite{li2019acoustic, gohil2018sentiment}, healthcare~\cite{gohil2018sentiment}, and public sectors~\cite{saxena2022unpacking, verma2022sentiment}, amongst other applications. Our finding from \minorediting{RQ1.a} implies that the extracted insights about subjectivity and polarity from textual data can vary significantly depending on which BSA tools are used.

Reading through the documentation, README files and associated research articles of our examined BSA tools indicated that none of these included post-development user testing and checking for identity-related biases. This leaves room for inconsistencies and discrepancies among sentiment analysis tools to go unscrutinized and unattended. Moreover, the lack of participation of users from different demographic groups within Bengali communities leads to disparities in accessing and using Bengali language technologies. Returning to our discussions on cultural hegemony in section~\ref{sec:literature_review}, such a digital divide among developers and users and invisible politics of code institutionalize a specific group's power and control through technological artifacts and, consequently, their perceptions and beliefs shape technology used within a larger community. By convincing others that their values and interests align with the overall community's perspectives and benefits, that specific group achieves technological hegemony. To resist certain groups systematically benefiting more from a sociotechnical system than other communities and systematically having influence over data-centric infrastructures, following prior scholarship~\cite{aragon2022human, amershi2014power}, we urge collaboration among stakeholders to ensure that their developed sentiment analysis tools' responses to Bengali sentences are aligned with the perspectives of the community and that they are not prejudiced against any particular identity or group of people.

\subsection{Codification of Implicit Expression of Identities in Sentiment Analysis}
To answer \minorediting{RQ1.b}, we examined how different BSA tools respond to different identity categories, expressed explicitly (e.g., through direct mention) and implicitly (e.g., through colloquial vocabularies, community norms around names and kinship). Similar to our examination of varied Bengali dialects in Bangladesh and India, other major languages have different dialects that are sociohistorically and culturally connected with particular groups within the broader linguistic communities (e.g., Southern and Coastal accents of American English, Quebec accent of French). For example, due to the refugee crises created by the postcolonial partition in Bengal, Bangladeshi (then East Bengal) dialects were associated with refugees in India, and speakers of this dialect are often subject to contempt both online and offline~\cite{das2021jol, chakrabarty1996remembered, chakrabarty2020migrant}. According to identity scholars, identity is constructed and learned through everyday speech acts and non-verbal activities in different social settings and are thus modeled after normative cultural and societal logics~\cite{butler2011gender}. Though researchers have qualitatively studied how sociotechnical platforms marginalize people based on their performative identity~\cite{das2021jol, scheuerman2019computers, munoz2022platform}, only a few works quantitatively studied how computing systems codify the performativity--the expression of identity through repetition of norms~\cite{butler2011gender} (e.g., colloquial verbal and speech acts) of various communities and groups~\cite{diaz2018addressing, sap2019risk}.

\editing{As parochial and stereotypical representations influence the development of datasets and tools, sentiment analysis and NLP tools broadly can inflict representational harm by conflating particular identities into one (e.g., viewing all Indic languages as the same or limiting a linguistic identity by nation states\footnote{\editing{Some decolonial scholars have argued that nation states and governments as forms of hierarchy and authority are also consequences of colonization that perpetuate colonial values (e.g. forced integration of smaller ethnic communities)~\cite{ramnath2012decolonizing, kapoor2012human}}}).} While researchers found evidence of accent gaps and racial disparity in speech recognition and language identifiers (e.g., not recognizing Southern American English)~\cite{blodgett2017racial, harwell2018why}, our study highlighted how sentiment analysis tools codify different country-based communities' preference of vocabularies as implicit expressions of identities and exhibit biases based on those. Prior CHI literature proposed using readily available sentiment analysis (e.g., VADER) to gather insights from textual data in algorithmic decision-making~\cite{saxena2022unpacking, rombergmaking}. Based on our finding that sentiment analysis tools codify the internal practices of different religion and nationality-based communities, we need to ask how these community practices and various societal biases and prejudices regarding those practices being embedded within sentiment analysis tools would impact algorithmic decision-making. We explore this issue further through the application of sentiment analysis tools in the context of content moderation in the following section.

\subsection{Exploring Downstream Effects of Bias in Sentiment Analysis Through the Context of Content Moderation}
In \minorediting{RQ2.a}, we found that most sentiment analysis tools available in the Bengali language are biased toward a particular category in cases of identity dimensions of gender, religion, and nationality. \editing{For sentences with similar structure and word content, most BSA tools (77\%) deemed Bangladeshi identity to be more positive than Indian identity, exhibiting a nationality-based bias. We found BSA tools exhibiting such favoritism toward female (38\%), male (30\%), Hindu (30\%), Muslim (38\%), and Indian (15\%) identities. Such preference toward a particular religious or national community's direct mention or linguistic practices resembles~\cite{das2021jol}'s finding of biases in content moderation.} For some BSA tools, we could not find evidence of those consistently assigning significantly different sentiment scores to different identity categories under a single dimension (e.g., T1 for gender, T5 for religion, and T10 for nationality). While those tools did not show bias in a particular dimension, our analysis could not identify a BSA tool that maintains such impartiality across all three dimensions of gender, religion, and nationality. Using biased language technologies like a sentiment analysis tool can have downstream effects. For example, sentiment analysis is also a ubiquitously used component in automated content moderation systems~\cite{vaidya2021conceptualizing, hettiachchi2019towards, smith2022instagram, sun2022design}. Scholarship in social computing and communications have studied the construction of automated content moderation~\cite{jhaver2019human, chandrasekharan2018internet} and users' perception of those systems~\cite{jhaver2019does, seering2019moderator}. Though, due to algorithms' complexities and common failure to understand the contexts of human languages, automated content moderation's legitimacy is questioned~\cite{pan2022comparing}, users perceive automated moderation to be more impartial with human oversight~\cite{ozanne2022shall}. Related to user personality and social aspects~\cite{riedl2022trust}, in some cases, researchers have found that ``users trust AI as much as humans for flagging problematic content"~\cite{molina2022ai, psu2022UsersTrust}.

Given how the transnational and religiously diverse Bengali communities' colonial past continues the distrust and division across religions and national borders and impacts their experience with platform governance and perception of biased content moderation, especially the anonymous human moderators~\cite{das2021jol}, we ask if automated content moderation is used instead of human moderation, how would that impact user interaction and experience for diverse Bengali communities? This question stems from considering ``automated" and ``human" as two ends of a spectrum of moderation style~\cite{jiang2023trade}. If the sentiment analysis component within that automated moderation system is biased, as we found in our study, it can misinterpret non-normative opinions as negative and trigger automated content moderation systems to remove the content from the platforms. Thus, users, especially the ones from marginalized and minority communities, can fear being censored for expressing their perspectives. Rather than complementing human moderators' efforts in managing large online communities, automated moderation can be employed as a pretext to justify the marginalization of diverse voices. Altogether, biased BSA tools being used in automated moderation can deter inclusive and in-depth discussions, prompt users to disengage or become inactive, and eventually shape a homogenized identity and reflect existing colonial divisions and structures in Bengali societies--much like the outcomes of biased human moderators~\cite{das2021jol}.

\subsection{Collaboration among Developers of Diverse Demographic Backgrounds}
Returning to \minorediting{RQ2.b}, though we did not find any relationship between the BSA tools' direction of biases and the demographics of those tools' developers, we cannot overlook the homogeneity of developers' identities. \editing{Since all the BSA tools we audited were developed by Bengali developers and not some Western entities, do we need to ask ``who designs?" Does postcolonial computing's concern about computing systems' similarities with colonial practices apply here? Prior CHI research found that while transgenerational colonial values (e.g., collective identity posited on difference) shape Bengali users' interaction with and through computing systems, collaborative discourses resist such views~\cite{das2022collaborative}. However,} earlier in the paper, in Table~\ref{tab:examined_tools}, we saw that most BSA tools on PyPI and GitHub are developed by solo developers or teams \minorediting{of a few coders with little diversity}--most tools being developed by individuals who identify as male, Muslim, and Bangladeshi. \editing{Similar to colonial Bengal, where certain exclusionary social identities (e.g., \textit{babu}: educated Bengali men often based in Kolkata, West Bengal) emerged as accepted changes in Bengali identity and subjectivity~\cite{dutta2021packing},} despite the Bengali language being spoken natively by diverse religious and national communities, we found certain isolation and lack of collaboration to exist among developers of diverse backgrounds. For example, though BSA \minorediting{tool T4} had both female and male developers, similar collaboration did not occur across various religion and nationality-based identities in any BSA tool.

Does the colonial past of the subcontinent and the Bengali people have anything to do with today's lack of collaboration in the developing sociotechnical systems in the Bengali context? Prior work has highlighted that colonial rule fragmented the Bengali people's imagination of communities, deepened the communal distrust among Hindus and Muslims, and increased the communication gaps among Bengalis in Bangladesh and India~\cite{chatterjee1993nation, das2021jol}. For example, whereas Indian Bengalis' nationality is shaped by linguistically diverse Indian identity~\cite{singh2018limits}, Bangladesh defines its people's concept of nationalism as being derived from Bengali language and culture~\cite{bangladesh1972constitution}. Therefore, language's role in shaping Bengali people's cultural identity and imagination of communities varies in Bangladesh and India. This difference translates to Bengali researchers' participation in computational linguistics research in their local language. For example, \minorediting{developers of all but one BSA tool self-identified as being} from Bangladesh. Beyond our study, most leading Bengali NLP research endeavors, such as learning and research groups\footnote{\url{https://bengali.ai/}, \url{https://csebuetnlp.github.io/}, \url{https://sustbanglaresearch.org/}} and workshops\footnote{\url{https://blp-workshop.github.io/}}, are supported and advanced by Bangladeshi communities and government. Though Indian researchers also regularly contribute to Bengali NLP, it is often done through the framing of NLP for Indic languages~\cite{arora2020inltk} and lacks the concentrated attention that the Bangladeshi NLP community puts in the Bengali language. As NLP tools in Bengali are predominantly developed by Bangladeshi Bengalis, those technologies, reflecting Bangladeshi values, norms, and prejudices, can become biased. Actively collaborating among individuals from different religions within the Bengali communities and institutions across geographic boundaries can contribute to mitigating such digital divisions.

%% file: sections/conclusion.tex
\section{Conclusion: Call for Engineering Activism in Critical HCI}
This paper presents findings from algorithmic audits of Bengali sentiment analysis (BSA) tools. Using statistical methods, we found that sentiment scores from different BSA tools vary for sentences with identical lexical content and structure. Our analysis also found evidence of BSA tools exhibiting biases, such as by consistently assigning significantly different sentiment scores to sentences expressing different gender, religion, and nationality-based identities. Complementing qualitative identity literature in CHI, we quantitatively examined how sentiment analysis tools respond to explicit and implicit expressions of a certain identity category in a sentence. In our discussion, we explained our quantitative findings through a postcolonial understanding of the studied linguistic communities' social, cultural, and historical contexts. Overall, this paper, foregrounding the historically marginalized and under-represented Bengali community, contributes to the intersection of CHI, social computing, NLP, and fairness and bias literature contextualized in the Global South.

While critical HCI studies adopting a qualitative approach can provide deep and rich insights into biases in computational systems, those explorations are insufficient and a fine-grained understanding of systems, architecture, algorithms, and code is essential for describing and explaining new information technologies' social, ethical, and political dimensions~\cite{nissenbaum2001computer}. Building on that call for ``engineering activism"--the use of engineering skills and knowledge to promote social justice, we argue that future NLP research (e.g., developing sentiment analysis tools), especially in critical HCI space, should actively reflect on identity-related biases and seek collaboration among individuals of diverse religious and transnational identities.